\documentclass[journal]{IEEEtran}
\usepackage[usenames]{color}
\usepackage{epsfig}
\usepackage{graphics}
\usepackage{amsmath}
\usepackage{amssymb}
\usepackage{multirow}
\usepackage{cite}
\usepackage{array}
\usepackage{pslatex} 
\usepackage{url}
\usepackage{caption}
\usepackage{lineno}
\usepackage{graphicx}  
\usepackage{setspace}
\usepackage{tikz}
\usepackage{hhline}
\usepackage{mathtools}

\usepackage{tabularx}
\usepackage{multirow}
\usepackage{booktabs}
\usepackage{etoolbox}
\newrobustcmd{\B}{\bfseries}
\usepackage{color}

\usepackage[letterpaper]{geometry}
 \newcommand{\flogo}{
 }
\ifCLASSINFOpdf
\else
\fi
\usepackage[english]{babel}
\usepackage[utf8]{inputenc}
\usepackage{fancyhdr}
\begin{document}
%
\title{\textcolor{black}{Preoperative Prognosis Assessment of Lumbar Spinal Surgery for Low Back Pain and Sciatica Patients based on Multimodalities and Multimodal Learning\vspace{0.25cm}}}

%
%
%

\author{Li-Chin Chen,
        Jung-Nien Lai,
        Hung-En Lin,
        Hsien-Te Chen, 
		Kuo-Hsuan Hung, 
		Yu Tsao,~\IEEEmembership{Senior~Member,~IEEE}
\thanks{This work was supported by the Ministry of Science and Technology in Taiwan under Grant MOST 110-2320-B-039-041-. Yu Tsao and Jung-Nien Lai contributed equally.
	\par Li-Chin Chen is with the Research Center for Information Technology Innovation, Academia Sinica, Taipei, Taiwan. Jung-Nien Lai is with the Department of Chinese Medicine, China Medical University Hospital, and School of Chinese Medicine, College of Chinese Medicine, China Medical University, Taichung City, Taiwan. Hung-En Lin is with the Department of Chinese Medicine, China Medical University Hospital, Taichung City, Taiwan. Hsien-Te Chen is with the Department of Orthopedic Surgery, China Medical University Hospital, and the Department of Sports Medicine, College of Health Care, China Medical University, Taichung City, Taiwan. Kuo-Hsuan Hung is with the Research Center for Information Technology Innovation, Academia Sinica, Taipei, Taiwan. *Yu Tsao is with the Research Center for Information Technology Innovation, Academia Sinica, Taipei, Taiwan (correspondence e-mail: yu.tsao@citi.sinica.edu.tw).
}}

\maketitle\thispagestyle{fancy}

\begin{abstract}
\textit{Goal:} Low back pain (LBP) and sciatica may require surgical therapy when they are symptomatic of severe pain. However, there is no effective measures to evaluate the surgical outcomes in advance. This work combined elements of Eastern medicine and machine learning, and developed a preoperative assessment tool to predict the prognosis of lumbar spinal surgery in LBP and sciatica patients. \textit{Methods:} Standard operative assessments, traditional Chinese medicine body constitution assessments, planned surgical approach, and vowel pronunciation recordings were collected and stored in different modalities. Our work provides insights into leveraging modality combinations, multimodals, and fusion strategies. The interpretability of models and correlations between modalities were also inspected. \textit{Results:} Based on the recruited 105 patients, we found that combining standard operative assessments, body constitution assessments, and planned surgical approach achieved the best performance in 0.81 accuracy. \textit{Conclusions}: Our approach is effective and can be widely applied in general practice due to simplicity and effective.
\end{abstract}


\IEEEpeerreviewmaketitle

\textbf{\textit{Impact Statement-} This research provides a light-weighted preoperative assessment tool that predict the prognosis of lumbar spinal surgery in low back pain and sciatica patients based on multimodalities and multimodel learning.}\\
\\

\section{INTRODUCTION}

\IEEEPARstart{L}{ow}  back pain (LBP) symbolizes pain and discomfort that is localized below the costal margin and above the inferior gluteal folds, having a lifetime prevalence of 60–80\% among the global population, making it one of the most common health complaints \cite{daniell2018failed}. Sciatica is neurological symptoms that involves the nerve root to be symptomatic, such as inflammation and compression \cite{valat2010sciatica}. Both LBP and sciatica are common symptoms that caused by spinal disorders, and may lead to severe pain. Pain severity may vary considerably among individuals \cite{long1996persistent}. 
Treatments for LBP and sciatica are similar and include non-invasive and invasive procedures. Most patients took the non-invasive approach to alleviate pain, including physical therapy and medication. Some patients receive more aggressive treatments, such as spinal surgery, intradiscal therapy, and narcotic and psychoactive drugs \cite{bernstein2017low,petersen1999classification,baron2004neuropathic}. However, none of these approaches guarantee symptoms elimination \cite{long1996persistent}. A significant number of lumbar postsurgical patients have been reported to continue to suffer persistent pain and limited function \cite{daniell2018failed, amirdelfan2017treatment}. The surgical outcomes were difficult to evaluate before surgery, and varies significantly between studies, known as the failed back surgery syndrome. Graz et al. \cite{graz2005prognosis} had indicated that while the physician foresees the patient to have “great improvement” before surgery, 39\% of the patients did not experience the least “minimal clinically important difference” in back pain after surgery. 

Ways of evaluating surgical outcomes have been proposed. Morlock et al. \cite{morlock2002nass} reported the combination of pain and function scale of patients before surgery can evaluate the success or failure of the surgery; Daltroy et al. proposed the North American Spine Society Lumbar Spine Outcome Assessment Instrument  \cite{daltroy1996north}. But none of them were effective, therefore, did not gain popularity. The complications of spinal surgeries can be severe and irreversible \cite{nasser2010complications}, which makes it a high risk treatment that requires comprehensive assessments before the intervention. 


Previous research extensively investigated the use of machine learning to extract feasible information from patient data \cite{won2020spinal,tougaccar2020application}. These data diverse in feature spaces and dimensionalities, but are also connected and have complex interactions in representing patients and their conditions. Based on this information, the medical staffs tailored a treatment plan. 
Integrating heterogeneous information better replicates the behavior of human experts, where multiple pieces of information are assembled before decisions are made \cite{huang2020fusion}. In machine learning, multimodal fusion learning is to combine data from different modalities and improve model performances \cite{huang2020fusion,atrey2010multimodal}. 

In this study, we bridge the gap through adding elements of Eastern medicine and machine learning technique, and developed a preoperative assessment tool based on multimodal learning to predict the prognosis of lumbar spinal surgery in patients with LBP and sciatica. We include modalities such as tabular data, free text, and audio recordings, and explored the effectiveness through comparing the performance of uni- and multimodal in different modalities, machine learning models, and combination of different fusion strategies. 

\subsection{Multimodal fusion in biomedical fields}
Data fusion refers to the process of joining data from multiple modalities, extracting feasible feature embeddings, and increasing the precision of predictions \cite{huang2020fusion}. Given two modalities, $X_{a}=X_{1}^{n}$,\ldots,$X_{t}^{n}$ and $X_{b}=X_{1}^{m}$,\ldots,$X_{t}^{m}$. $X_{t}^{n}$ and $X_{t}^{m}$ refer to the $\emph{n}^{th}$ and $\emph{m}^{th}$ dimensional feature vectors of the $X_{a}$ and $X_{b}$ at time \emph{t}, where $X_{a}^{T}\in\mathbb{R}^{i}$ and $X_{b}^{T}\in\mathbb{R}^{j}$, \emph{i} and \emph{j} represent the different times and spaces of each modality, and $\emph{T}=\{1$, \ldots, $\emph{t}$\}. Given the ground truth labels $\emph{Z}=Z^{1}$, \ldots, $\emph{Z}^{t}$, we train a multimodal learning model \emph{M} that maps $X_{a}$ and $X_{b}$ into \emph{Z}. Using $N_{a}$ and $N_{b}$ to denote the unimodal networks from $X_{a}$ and $X_{b}$, respectively. $N_{a}: X_{a} \rightarrow \emph{Y}$, $N_{b}: X_{b} \rightarrow \emph{Y}$, and $M= N_{a} \bigoplus N_{b}$. \emph{Y} denotes the predicted class label generated by the output of $N_{a}$ and $N_{b}$, $\bigoplus$ indicates the concatenation operation, and $M$ symbolized the generated multimodal network that learned from the concatenated representation from different modalities \cite{bayoudh2021survey}.

Based on fusion at different level, there are three commonly known data fusion strategies: Early fusion (EF) refers to joining multiple input modalities into a single feature vector before feeding into the machine learning model, it utilized the correlation between different modalities at an early stage; Joint fusion (JF) is the process of joining the learned feature representation of multiple modalities from different models, and fusion the joined representation with an additional model to make the prediction. The loss of the final prediction model backpropagated to the feature extracting neural networks, creating better feature representations for each modality \cite{huang2020fusion}; Finally, late fusion (LF) combined the prediction results of multiple models, and use aggregated units to reach the final decision. Merely combining the results of multiple machine learning algorithms makes the assembling easier, not limited to neural networks (NN) \cite{huang2020fusion,atrey2010multimodal}.

Multimodal fusion is commonly used when attempting to combine information from heterogeneous data types. Venugopalan et al. \cite{venugopalan2021multimodal} integrated magnetic resonance imaging (MRI) imaging, single nucleotide polymorphisms (SNPs), and clinical test data to classify patients with Alzheimer’s disease; Lai et al. \cite{lai2020overall} used microarray and clinical data to predict the occurrence of non-small cell lung cancer. Other modalities, such as the electrocardiogram (ECG), computed tomography (CT), single-photon emission computerized tomography (SPECT), positron emission tomography (PET), clinical data, and clinical text \cite{zhou2021radfusion, li2021medical}, were used to support clinical practices. 
Despite the prosper investigation of machine learning applications in biomedical field, insufficient work in exploring the combination of different modalities, multimodal, and fusion strategies in preoperative assessments had been indicated \cite{huang2020fusion}.

\section{MATERIALS AND METHODS }

In this study, we aimed to develop a preoperative assessment tool that accurately assessed the prognosis of lumbar spinal surgery in patients with LBP and sciatica before surgical therapy. This study used a prospective cohort approach in which patients were recruited and assessed before and after surgery. This study was approved by the institutional review board (IRB) of China Medical University Hospital, Taichung, Taiwan (No. CMUH109-REC3-094).

\subsection{Patient recruitment}
From July, 2020 to July, 2022, patients diagnosed with LBP or sciatica, over the age of 12 years, with clear consciousness and capability in verbal communication, were considered as candidates for this study. Patients who had been treated in the outpatient setting of Chinese medical departments, China Medical University Hospital, Taiwan, and due to disc herniation, spinal stenosis, spondylolisthesis, or degenerative disc disease that were considered to pursue surgical treatments were approached and recruited after informed consent was obtained. Patients who had undergone spinal surgery before, with an American Society of Anesthesiologists (ASA) classification higher than IV, or diagnosed with any psychological disorder, dementia, drug addiction, or any disease that could potentially affect their linguistic ability or verbal pronunciation, were excluded. Patients were assessed with three types of assessments, namely the standard operative assessments, traditional Chinese medicine (TCM) body constitution assessments, and recordings of vowel pronunciation. The assessments were done and scored by trained clinical staffs to maintain maximum objectiveness. All surgeries were performed by the same orthopedic surgeon with more than 30 years of experience. Assessments were evaluated twice, before the surgery and during the first outpatient follow-up visit after discharge, respectively.

\subsection{Standard operative assessments and planned surgical approach}
Before spinal surgery, demographic information such as age, sex, body mass index (BMI), and four standard operative assessments were evaluated. The four standard assessments are the Visual Analog Scale (VAS), EuroQol Five Dimensions (EQ-5D), Oswestry Disability Index (ODI), and American Society of Anesthesiologists (ASA) \cite{gender_spine_surgery,delgado2018validation,poder2020predicting,doyle2017american,fairbank2000oswestry,haefeli2006pain,herdman2011development}, each of which represents the degree of pain, quality of life, level of disability, and level of risks of the operation. 
VAS evaluates the degree of pain that a patient feels ranging from 0 to 10; EQ-5D calculates the utility scores and the quality-adjusted life-year ranging from 0 to 1.000, assessing the number to the third decimal point; ODI is specified in measuring the disability in daily life related to LBP ranging from 0 to 100\% in the form of percentage; and finally, ASA is a simple categorization of patient physiological status that predicts the risks of the operation. 

Planned surgical approach included the information of the spinal segment of the operation and the intended intervention method. Surgeries include five different approaches, namely the minimally invasive surgical transforaminal lumbar interbody fusion, percutaneous endoscopic lumbar disc discotomy, percutaneous endoscopic discectomy and drainage, transforaminal lumbar interbody fusion, and anterior cervical discectomy and fusion. This information was written in the form of free text summary. 

\subsection{Traditional Chinese medicine assessment}
In TCM, \emph{Yin} and \emph{Yang} represent the energy balance of an individual. When \emph{Yin-Yang} is either imbalanced (Yin deficiency (\emph{Yin-Xu}) or Yang deficiency (\emph{Yang-Xu})) or loses its harmonizing dynamic (\emph{Stasis}), physical symptoms and signs appear. \emph{Stasis} represents the attenuation or obstruction of the energy flow in the body, referred to as the dynamic interaction between \emph{Yin} and \emph{Yang}, is slowed down and less efficient, and may express some physical symptoms, such as dizziness, chest tightness, or numbness in the limbs. When a patient does not possess the constitution of \emph{Yang-Xu}, \emph{Yin-Xu}, or \emph{Stasis}, the patient is considered to have a \emph{gentleness} constitution. \emph{Gentleness} describes a balanced state of qi, blood, and energy in the body \cite{lin2012bcqs}.

This study adopted the Body Constitution Questionnaire (BCQ) \cite{lin2012bcqs} to assess individual differences among patients. The questionnaire consisted of three sections, 44 questions that were based on a 5-point frequency scale (1 to 5, from “never happened” to “always the case”). 
A patient was diagnosed with \emph{Yang-Xu} when the patient scored 31 points or above in the \emph{Yang-Xu} section. Similarly, the thresholds for \emph{Yin-Xu} and \emph{Stasis} are 30 and 27, respectively. A patient can be diagnosed with multiple constitutions simultaneously, and those who were not diagnosed with any constitution were labeled as \emph{gentleness}. 

In this study, the TCM body constitution acted as seven input features to our model. We used the individual scoring of each constitution (\emph{Yang-Xu score}, \emph{Yin-Xu score}, and \emph{Stasis score} as numerical features), and the diagnosed result of the body constitution based on the mentioned thresholds (Yes/No as categorical features). The diagnosed results were added to minimize the disturbances of long questionnaire when patients become tiresome or subjective responses during the assessment. Finally, \emph{Gentleness} will be labeled as “Yes” for those with all three constitutions below the threshold, and “No” for those diagnosed with any constitutions (categorical feature).

\subsection{Vowel pronunciation}

Vowels have been considered an articulatory basis and have a distinct role in phonology \cite{lin2012bcqs}.
They are an essential component of language pronunciation. From the TCM perspective, vowel pronunciation symbolizes the imbalance of inner energy in an individual \cite{chiu2000objective}. We recorded five vowel utterances of each patient (including pronouncing /a/, /e/, /i/, /o/, and /u/) to represent their acoustic features. Each subject produced sustained stable phonation three times at a comfortable pace with no prior coaching or training. Patient voices were recorded using a commercial microphone (Shure SM58). The microphone was set at a distance of 15-20 cm from the patient. All voice samples were digitized using a 16-bit analog-to-digital converter at a 44.1 kHz sampling rate. The microphone was connected to a computer, and the audio file was stored in .mp3 format. 

\subsection{Prognosis determination}
\begin{figure}
\centerline{\includegraphics[width=\columnwidth]{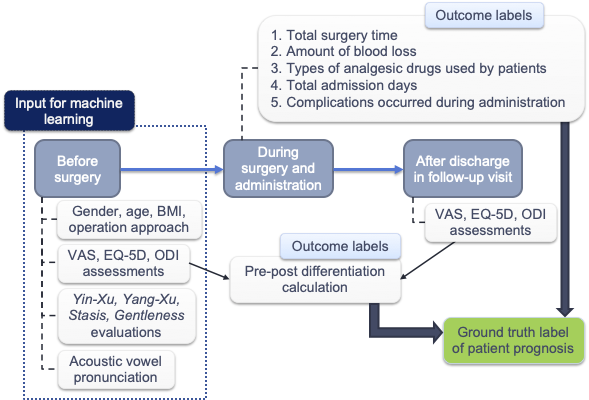}}
\caption{Patient assessments and ground truth determinations. BMI: Body mass index; VAS: Visual Analog Scale; EQ-5D: EuroQol Five Dimensions; ODI: Oswestry Disability Index.}
\label{fig_ground_truth_determ}
\end{figure}

The prognosis of patients was categorized into a binary label, desirable and undesirable, and served as the ground truth of this work. To evaluate the prognosis of patients, patient status before, during, and after the surgery was assessed. Combining the pre-post differentiation of VAS, EQ-5D, and ODI, status during the surgery, including the total surgery time, amount of blood loss, types of analgesic drugs used by patients, total admission days, and whether complications occurred during the administration, a total of eight outcome scores were used. 
The mean value of each scores was calculated and used as a threshold. Patient scored higher than average were labeled as desirable, while those scored lower than average were labeled as undesirable. With the eight outcome labels, the average number of desirable labels was calculated. If a patient gained more desirable labels than average, then the prognosis was determined as desirable; if a patient gained desirable labels less than average, then the prognosis would be categorized as undesirable. The final determination of patient prognosis was the ground-truth label used in this study. The determination flow is shown in Fig.~\ref{fig_ground_truth_determ}.

We intend to develop a prognosis assessment tool that operates as a reference when physicians decide patient treatments. Based on our model, physicians can make more precise foresight of how patients respond after surgery. Therefore, merely information obtained before surgery was included in the model training process. Information acquired during and after surgery was used to generate the prediction labels and validate the prediction results.

\subsection{Uni- and multimodal fusion design}
Owing to the requirement of combining different modalities in this research, we designed the model based on different data types, namely tabular data, free text, and acoustic recordings. 
When using single modality in the unimodal design, each unimodal design consisted of a preprocessor $Pre$ that preprocessed the data, an encoder $En$ that extracted feasible features, and a classifier $C$ that reached the final prediction. Their differences were denoted in subscripts.  

\begin{figure}
\centerline{\includegraphics[width=0.8\columnwidth]{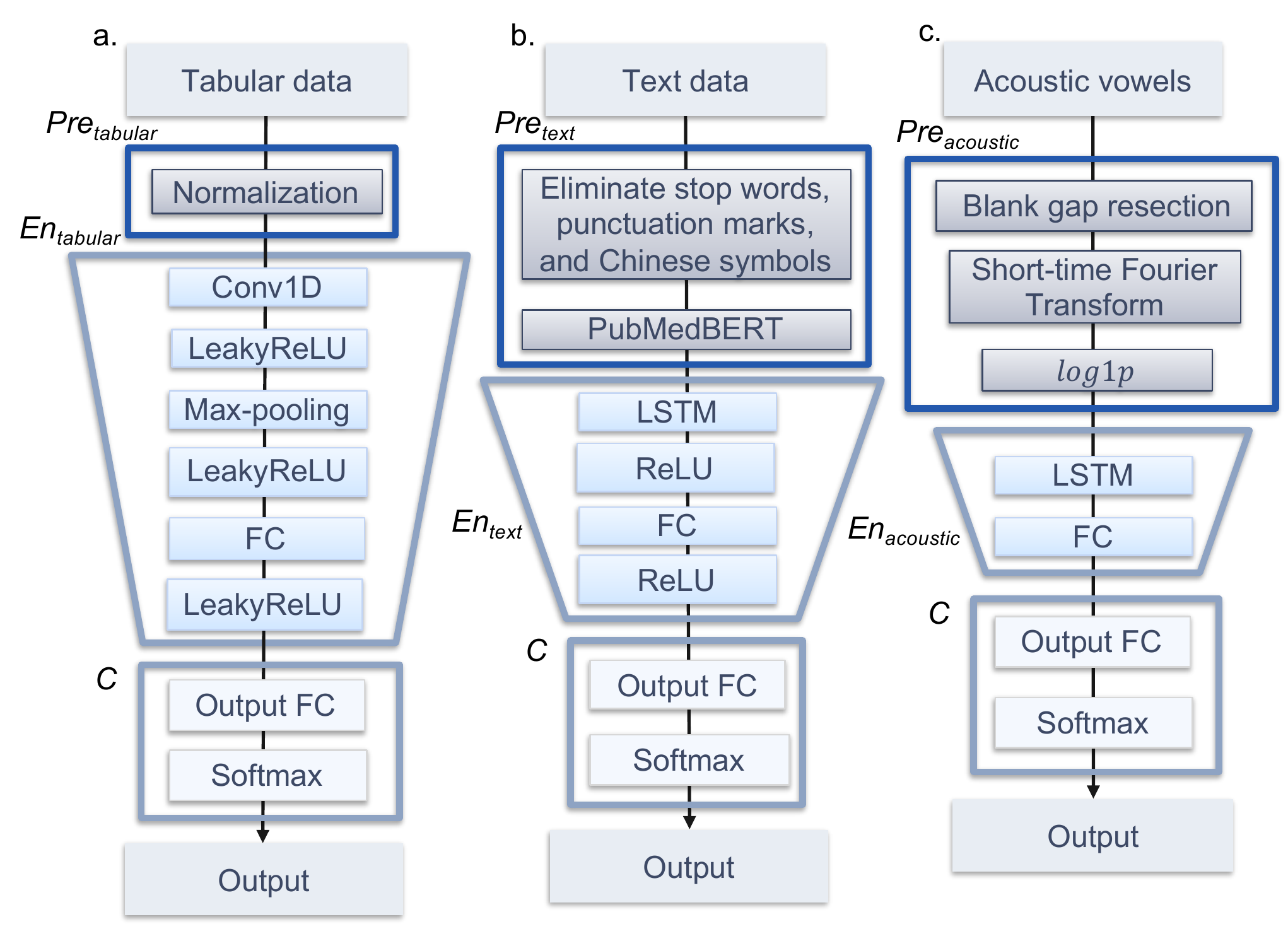}}
\caption{Architecture design of uni-model for each modality. Conv1D: one-dimensional convolutional neural network; LeakyReLU: leaky rectified linear unit; FC: fully connected; LSTM: long short-term memory; $Pre$: preprocessor; $En$: encoder; $C$: classifier;  $log1p$: natural logarithm of one plus the input.}
\label{fig2}
\end{figure}

\subsubsection{Tabular data}
In this study, the preprocessor $Pre_{tabular}$ scaled each numerical value between -1 and 1, and the categorical features were transformed into one-hot encodings. Both are then concatenated into a vector to represent each patient. Research have shown that treating the tabular features as an image-like vector and use convolutional layers can be beneficial \cite{diabetes_JBHI,CNN_tabular}. Hence, we used a convolutional neural networks (CNN)-based encoder $En_{tabular}$, which includes the combination of one-dimensional convolutional (Conv1D) layers, fully connected (FC) layers, leaky rectified linear unit (ReLU) activation functions, and max-pooling layers, shown in Fig.~\ref{fig2}a. The convolution kernel size was set to three, stride was two, and the width of the max-pooling was three. The FC layer had a single hidden layer with 20 neurons followed by a ReLU function. The final classifier $C$ consisted a FC output layer that have two output nodes to condense the result to a binary output, and followed by a softmax activation function to normalize the output to a probability distribution that summed up to 1. Symbolizing the tabular modality as $X_{tabular}$, the process can be formulated as: 
\begin{equation} 
    \hat{y} = C[En_{tabular}(Pre_{tabular}(X_{tabular}))].
    \label{eq1}
\end{equation} 


Meanwhile, tree-based algorithms were well-known to achieve many state-of-the-art results in tabular data \cite{boost_compare}. We further tested several tree-based algorithms, and select the best performed one to perform multimodal fusion. The tree-based algorithms include Random Forest, Adaptive Boosting (AdaBoost), Categorical Boosting (CatBoost), eXtreme Gradient Boosting (XGBoost) , and Light Gradient Boosting Machine (LightGBM).


\subsubsection{Free text data}
Currently, PubMedBERT \cite{Gu2021} outperformed other language representation models in biomedical-specific natural language processing (NLP). Therefore, PubMedBERT was used as our representation extraction model for free text modality. The preprocessor $Pre_{text}$ first eliminated stop words, punctuation marks, and Chinese symbols in the writings, and used PubMedBERT as a frozen pretrained model and transformed the writings into latent representations. Afterward, the encoder $En_{text}$ fine-tuned the results to the downstream task in this work. $En_{text}$ included a long short-term memory (LSTM) layer and an FC layer that both consist 20 hidden nodes.
The classifier $C$ finally output the prediction, shown in Fig.~\ref{fig2}b, and the process can be denoted as: 
\begin{equation} 
    \hat{y} = C[En_{text}(Pre_{text}(X_{text}))],
    \label{eq7}
\end{equation} where $X_{text}$ represented the free text modality. 

Meanwhile, tree-based algorithms were also tested, and the best performed one were selected to do further fusion design.

\subsubsection{Audio vowel recordings}

In the audio files, the blank gap between each pronunciation was resected to ensure that all the audio sounds consisted of the patients’ voices. The files were then transformed into a waveform audio file format (.wav), and processed with Short-time Fourier transform (STFT) to a spectrogram. The natural logarithm of one plus the input ($log1p$) was then applied to ensure the value is above zero, which normalized the potential error that could have contributed by the maldistribution between the positive and negative values. 
Mel-frequency cepstrum (MFCC) was also generated. The best performed feature was chosen for further fusion design. These preprocessing were all included in the preprocessor $Pre_{acoustic}$.

We treat each utterance (patient pronouncing /a/, /e/, /i/, /o/, and /u/) as a singular record that predicts patient prognosis; that is, assuming that the number of recruited patients was $p$, the model was trained on $p \times 5$ records. Two audio files from the same patient with poor quality were excluded, leaving 523 files entered the analysis, with framing set to 256. Considering that the audio signal also consists of time-series information, the encoder $En_{acoustic}$ also used an LSTM layer and an FC layer to extract sequential features. The LSTM layer consisted of seven hidden nodes, and the FC layer consisted of 10 hidden nodes, followed by the classifier $C$. The architectural design is shown in Fig.~\ref{fig2}c. Denoting the acoustic modality as $X_{acoustic}$, the process can be represented as: 
\begin{equation} 
    \hat{y} = C[En_{acoustic}(Pre_{acoustic}(X_{acoustic}))].
    \label{eq8}
\end{equation}

The audio recordings possess high dimensionality, which is required to be condensed before entering tree-based algorithms. This would make the comparison difficult, unable to determine whether the results were affected by the input features or the model designs. Our work focused on the model design, as a result, tree-based algorithms were not tested on audio modality. 

\subsubsection{Multimodal fusion design}
We examine the EF, JF, and LF strategies in multimodal fusion. All modalities were processed with preprocessors in advance. NN and the chosen tree-based algorithm were tested, and combined with different fusion strategies. First, the effect of using a single modality in predicting the patient prognosis (tabular data, free text data, and acoustic vowels) was demonstrated, then two modalities (tabular + text, tabular + audio, text + audio), and finally three modalities (tabular + text + audio). If the combination required audio recordings, then LF would be performed to fuse heterogeneous model results.

\begin{figure}
\centerline{\includegraphics[width=0.6\columnwidth]{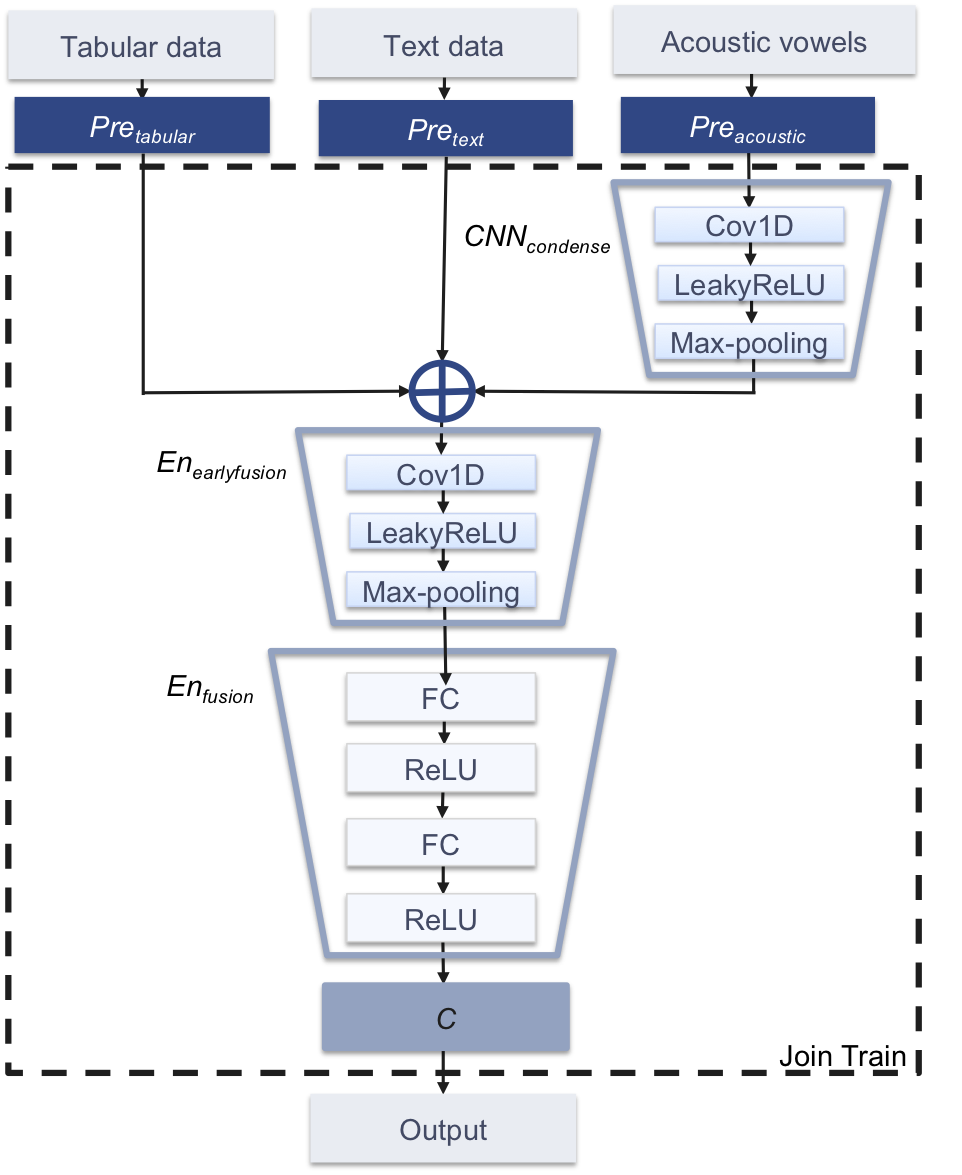}}
\caption{Architecture of multimodal in early fusion design. $Pre$: preprocessor; $CNN$: convolution layer; Conv1D: one-dimensional convolutional neural network; LeakyReLU: leaky rectified linear unit; $En$: encoder; FC: fully connected; ReLU: rectified linear unit; $C$: classifier; $\bigoplus$: concatenation operation.}
\label{fig_EF}
\end{figure}


\begin{figure}
\centerline{\includegraphics[width=0.6\columnwidth]{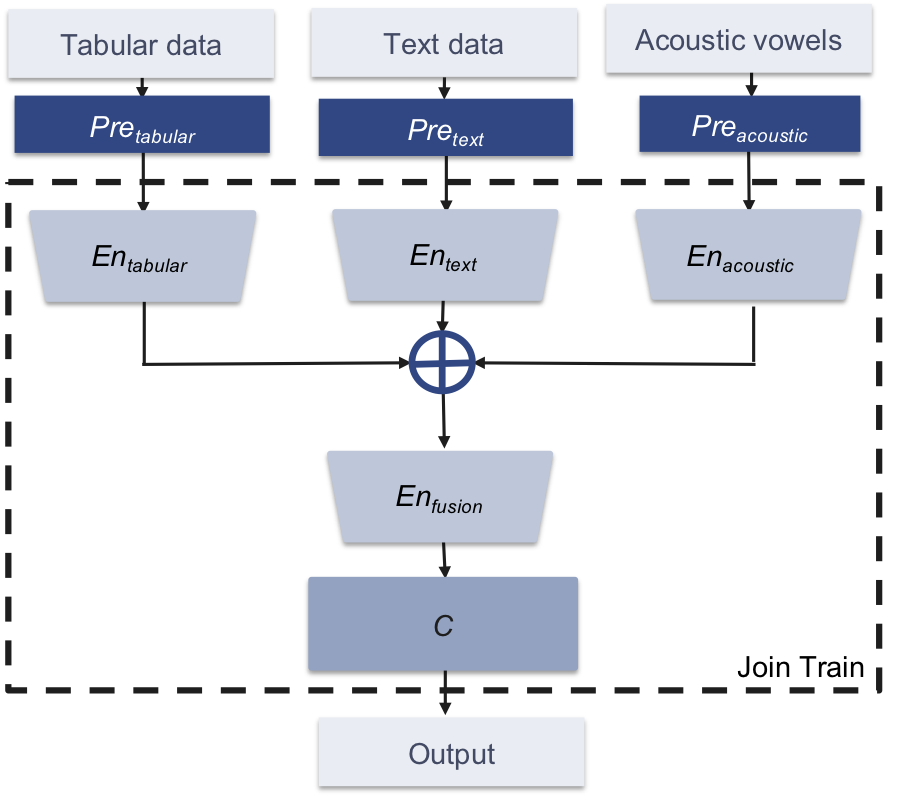}}
\caption{Architecture of proposed multimodal in joint fusion design. $Pre$: preprocessor; $En$: encoder; $C$: classifier; $\bigoplus$: concatenation operation.}
\label{fig_JF}
\end{figure}

In EF design, merely the audio modality was first condensed with a $CNN_{condense}$ layer to enable the concatenation with other modalities. The $CNN_{condense}$ layer was joined trained with the rest of the model, which ensured the dimensionalities were condensed through learned parameters. The concatenated vectors were also treated as an image-like vector, which was applied with a CNN-based encoder $En_{earlyfusion}$. $CNN_{condense}$ and $En_{earlyfusion}$ both consisted a Conv1D layer, a leaky ReLU function, and a max-pooling layer. Both convolution kernel size was set to three, stride was two, and the width of the max-pooling was three. The EF design of triple modalities is illustrated in Fig.~\ref{fig_EF}, and the fusion process can be expressed as:  
\begin{equation}
    \begin{aligned}
        \hat{y} = C\{En_{fusion}[&En_{earlyfusion}[\\
          & Pre_{tabular}(X_{tabular}); \\
          & Pre_{text}(X_{text}); \\
          & CNN_{condense}(Pre_{acoustic}(X_{acoustic}))]]\}.
        \label{eq_early_fusion}
    \end{aligned}
\end{equation}

The EF model using NN was trained and reach optimization based on the combination of triple modalities. Afterwards, the modalities were withdrawn sequentially to observe the contribution of each modality. In two-modality designs, tabular + text were tested with the chosen tree-based algorithm in the EF fusion process, other combinations (tabular + audio, text + audio) used NN-based fusion design. 

In the JF design, the modalities were concatenated after individual encoder extracted feasible embeddings. We used the same architecture of the best performing unimodal of each modality, removed the final classifier $C$, 
concatenated the latent features of different modalities, and fused it with encoder $En_{fusion}$, and finally classify the result with $C$. 
The detailed model design of triple modalities is illustrated in Fig.~\ref{fig_JF}, and the fusion process can be denoted as: 
\begin{equation}
    \begin{aligned}
        \hat{y} = C\{ En_{fusion}[ & En_{tabular}(Pre_{tabular}(X_{tabular})); \\
          & En_{text}(Pre_{text}(X_{text})); \\
          & En_{acoustic}(Pre_{acoustic}(X_{acoustic}))]\}.
        \label{eq9}
    \end{aligned}
\end{equation}
In two-modality designs, the modalities were also withdrawn sequentially. Tree-based algorithms does not generate latent features, hence, JF design can only be achieved through NN. 

\begin{figure}
\centerline{\includegraphics[width=0.6\columnwidth]{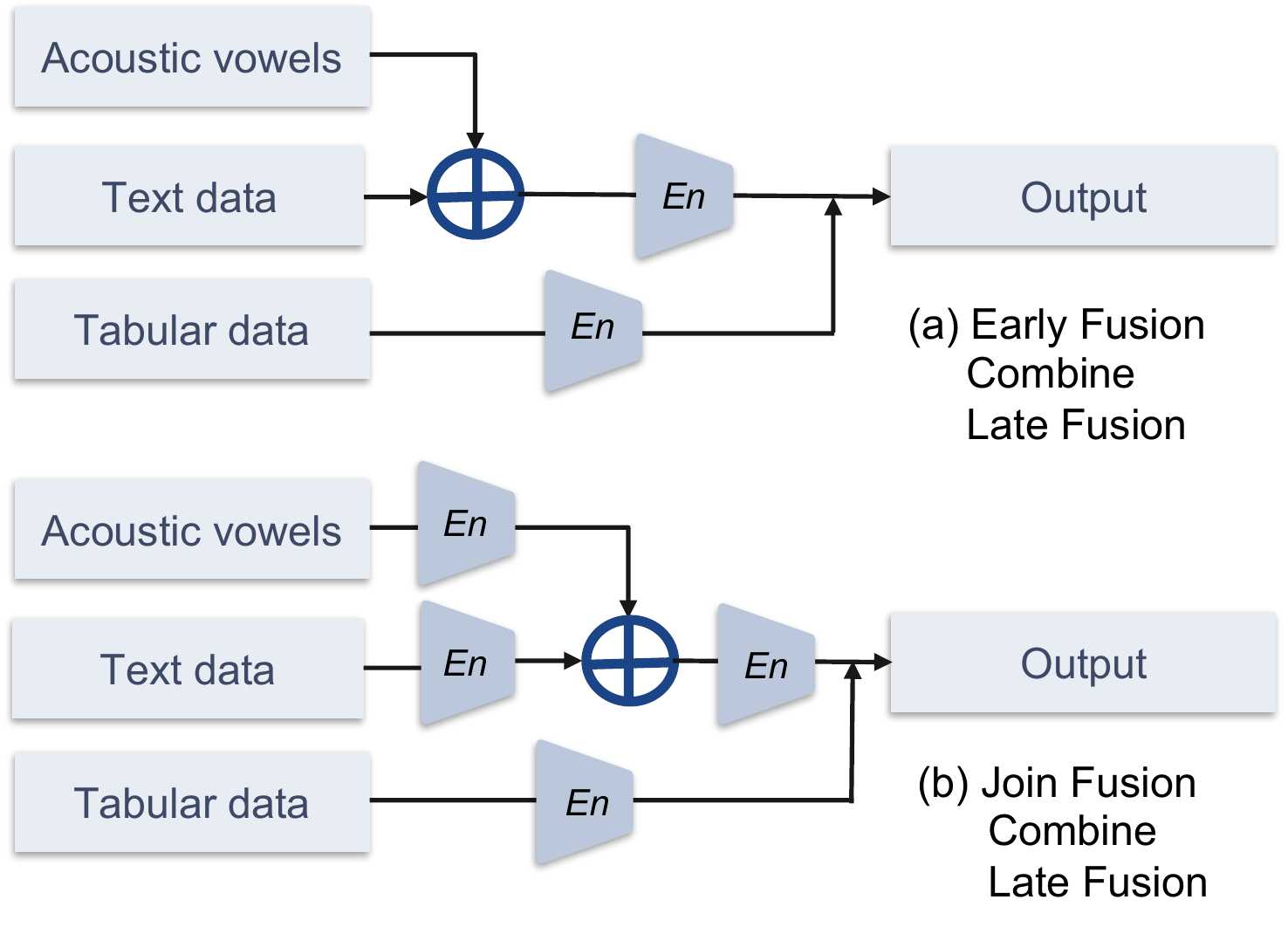}}
\caption{Late fusion design for triple modalities. $En$: encoder, refers to any kind of algorithm; $\bigoplus$: concatenation operation}
\label{fig_FusionStrategy}
\end{figure}

Finally, LF design was to integrate the result of NN and non-NN. In triple-modality fusion, the best performed model structure in unimodal design were chosen. After two models arrived at their predicted probabilities, two results were summarized based on a weighting percentage of 60:40 to reach the final conclusion. The better performed model received a higher weight. Due to the fact that tabular data outperformed in tree-based algorithm, leading the LF triple-modality fusion in concatenating text and audio modalities with NN, using EF and JF, respectively, and fused with tree-based probabilities of tabular data. This results in hybrid modes such as EF + LF and JF + LF, shown as Fig.~\ref{fig_FusionStrategy}a and b. 
In summary, the two-modality fusion were examined with EF, JF, and LF, and the the triple-modality fusion were tested with EF, JF, EF + LF, and JF + LF, respectively.

\subsection{Training process and performance evaluation}
To demonstrate the effect of using multiple modalities, we compare the differences among uni- and multimodal, different modality combinations, and fusion strategies. The aforementioned model designs were trained and optimized by iterating through small subsets of the training data and modifying the parameters to minimize the loss between the prediction and the prediction target. We adopted the five-fold cross validation strategy and L2 regularization. Also, to avoid data leaking problem, which the model achieves good performance by remembering the character of patients (such as gender) rather than the physiological features within the voices, each patient was assigned only to the training dataset or testing dataset. Thus, we ensured that the model was tested on utterly unseen individuals, which assured the credibility of the prediction model. During the model training process, the patients were randomly split into training and testing data at an 80:20 ratio. The training data were then randomly separated into five subsets. The training process included the rotation of each subset as the validation subset, whereas the others acted as training subsets. The reported performances was tested based on the testing data that was isolated in the beginning, which were unseen by the model and demonstrated the genuine performance. 


The evaluation metrics included the area under the receiver operating characteristic (AUROC) curve, accuracy, sensitivity, specificity, precision, and F1 score. To monitor the robustness of the model, that is, the ability to replicate the same result, two approaches were used. First, the reported metrics were the mean values of executing the training and testing process ten times. We considered the averaged performances of the testing samples to be more representative than the singular validation results. Second, we quantified the variation of the metrics by performing the non-parametric bootstrap approach to identify the 95\% confidence interval (CI) \cite{Briggs1998,Hazra2017}
indicating the boundaries of possible values under a 95\% confidence. The variation $v=|upperbond-lowerbond|$ of all metrics were calculated and averaged, resulting in a variation mean score (VMS) to represent the generalizability and stability of the model. 

\subsection{Interrelationship and interpretability}
To further explore the inner and outer correlation between the modalities, we adopted the Pearson correlation ($r$) test to analyzed the linear correlations between tabular features against the prediction label, and canonical correlation analysis (CCA) \cite{CCA} to analyzed the linear correlation between modalities. Whereas $r$ is a common analysis that demonstrates the linear correlation between variables; CCA is a technique that identifies a pattern that project the two multivariate datasets into sequence of paired scores (known as the canonical variates, can be seen as a type of latent variable), and the two scores exhibit maximum $r$. The pattern can be referred to as a linear regression of the variables within each dataset, and the $weights$ in CCA attributed to the coefficients of regressions, that indicates the variation of canonical variate when one unit of variable increased (when all other variable held constant). Meanwhile, the $loadings$ indicates the correlations between the variables and the canonical variate, symbolizing the contribution of the variable in the regressions. We took the first canonical variate that indicates the highest $r$ between the canonical variate of the other dataset. Due to the input elements in text and audio does not correspond to variables like tabular data that inherit distinct meanings, we observed the $loadings$ of each tabular variable between text modality and audio modality, respectively. 

Further, to increase the interpretability of our model, we adopted the Integrated Gradients (IG) \cite{captum_IG} and feature importance to analyze the decision making of the proposed model. IG computes the gradients of the prediction output to its input features, indicating an $attribute$ that refers to the contribution of the inputs to the prediction results. The computation can be represented as:

\begin{equation}
\begin{aligned}
Int&egratedGrads_{i}(x)\\
&::= (x_{i}-x_{i}^{\prime}) \times \int_{\infty=0}^{1} \frac{\partial F(x^{\prime}+\alpha \times (x-x^{\prime}))}{\partial x_{i}} d\alpha
\label{eq_IG}
\end{aligned}
\end{equation}
where $F: R^{n} \rightarrow [0,1]$ represents the neural network, $x \in R^{n}$ and $x^{\prime} \in R^{n}$ represent the input and baseline input (which is a zero embedded vector). IG computed the gradients at all points along the straightline path from $x^{\prime}$ to $x$, and $\frac{\partial F(x)}{\partial x_{i}}$ is the gradient of $F$ at the $i^{th}$ dimension. While the input size of text and audio largely exceed tabular data, the number of $attribute$ differs significantly. This made the sum or mean of $attribute$ of each modalities highly skewed. Thus, we compare the values and the distributions of $attribute$ among modalities. The interpretability of tree-based algorithms were explored through the measurement of feature importance. The most valued features in tabular data and the most valued modalities were discussed. 

The condensing trend of the training loss were demonstrated to further understand the convergence behavior of NN in EF and JF. Owning to LF were trained based on different losses and concatenate based on final prediction probabilities, the convergence of LF is not representative using NN loss trend, therefore, was not compared.

\section{RESULTS}

A total of 105 patients were recruited in this study. The patients’ demographic information, assessment results, and prognosis results are shown in Table~\ref{table_demographic}. It can be observed that more men were recruited (53\%), and the average age of patients was at the age of 56.76. The average number of desirable labels was 4.77, and was set as the threshold to categorize desirable and undesirable prognoses. Further patient information used to determine patient prognosis are shown in Table~\ref{appendix_table_demographic} in Appendix. 

\begin{table}
\caption{Demographic information of the recruit patients}
\label{table_demographic}
\setlength{\tabcolsep}{3pt}
\begin{tabular}{p{55pt}p{85pt}p{50pt}p{50pt}}
\toprule
\multicolumn{2}{c}{Items} & \multicolumn{2}{c}{Values}\\
\midrule
\multirow{8}{\linewidth}{Operative Assessment (preoperative)} & 
\multirow{2}{*}{Gender (n, \%)}& Male &56 (0.533) \\
                       && Female & 49 (0.467)\\
          & Age (mean, SD) & 56.757 & 16.423 \\
          & BMI (mean, SD) & 25.120	& 3.491  \\
          & VAS (mean, SD) & 4.467 & 3.009 \\
          & EQ-5D (mean, SD) & 0.612 & 0.150  \\
          & ODI (mean, SD) & 0.449 & 0.225  \\
          & ASA (mean, SD) & 2.219 & 0.537\\
\midrule
\multirow{5}{\linewidth}{TCM Body Composition Assessment (preoperative)} 
          & \emph{Yang-Xu score} (mean, SD) & 32.476	& 9.910 \\
          & \multirow{2}{*}{\emph{Yang-Xu} (n, \%)} & Yes & 50 (0.476) \\
                                                        && No & 55 (0.524) \\
          & \emph{Yin-Xu score} (mean, SD) & 31.267	& 9.478  \\
          & \multirow{2}{*}{\emph{Yin-Xu} (n, \%)} & Yes & 49 (0.467) \\
                                                      && No & 56 (0.533) \\
          & \emph{Stasis score} (mean, SD) & 26.857	& 9.387 \\
          & \multirow{2}{*}{\emph{Stasis} (n, \%)} & Yes & 28 (0.267) \\
                                                      && No & 77 (0.733) \\
          & \multirow{2}{*}{\emph{Gentleness} (n, \%)} & Yes & 45 (0.429) \\
                                                       && No & 60 (0.571) \\
\midrule  
\multirow{2}{\linewidth}{Ground Truth Label} & \multirow{2}{\linewidth}{Prognosis evaluation of patient, n (\%)}
               & Desirable  & 65 (0.619) \\
              && Undesirable & 45 (0.429) \\
\bottomrule
\multicolumn{4}{p{251pt}}{n: number of samples; SD: Standard deviation; BMI: Body mass index; VAS: Visual Analog Scale; EQ-5D: EuroQol Five Dimensions; ODI: Oswestry Disability Index; ASA: American Society of Anesthesiologists; TCM: Traditional Chinese medicine.} \\
\end{tabular}
\end{table}
\begin{table}
\centering
    \caption{Performance for singular modality in unimodal design}
    \label{table_singular}
    \begin{tabularx}{\columnwidth}{p{28pt}>{\centering\arraybackslash}p{60pt}>{\centering\arraybackslash}p{55pt}>{\centering\arraybackslash}p{55pt}}
    \toprule
    & Tabular data \break (Cat) & Free text \break (NN) & Audio vowels (STFT)\\
    \midrule
    AUROC       & 0.815 & \B 0.845 & 0.599  \\
    Accuracy    & 0.738 & \B 0.762 & 0.607  \\
    Sensitivity & 0.792 & \B 0.850 & 0.613  \\
    Specificity & \B 0.667 & 0.644 & 0.598\\
    Precision   & 0.760 & \B 0.768 & 0.672  \\
    F1          & 0.775 & \B 0.801 & 0.640  \\
    VMS         & \B 0.035 & 0.210 & 0.160 \\
    \bottomrule
    \multicolumn{4}{p{240pt}}{Cat: CatBoost; NN: neural networks; STFT: Short-time Fourier transform; AUROC: area under the receiver operating characteristics curve.}\\
    \end{tabularx}
\end{table}
\begin{table}
    \caption{Performance for different modality combinations in early fusion and joint fusion design using neural networks}
    \label{table_EF_JF}
    \centering
    \begin{tabularx}{\columnwidth}{p{5pt}p{32pt}>{\centering\arraybackslash}p{25pt}>{\centering\arraybackslash}p{25pt}>{\centering\arraybackslash}p{25pt}>{\centering\arraybackslash}p{35pt}>{\centering\arraybackslash}p{35pt}}
    \toprule
    &  &Tabular + \break Text & Tabular + Audio & Text \break + Audio & Tabular + Text + Audio & Tabular(LF) + Text + Audio\\
    \midrule
    \multirow{7}{\linewidth}{EF} 
    & AUROC       & \B 0.829 & 0.595 & 0.776 & 0.743 & 0.792 \\
    & Accuracy    & 0.733 & 0.630 & 0.734 & 0.706 & 0.731 \\
    & Sensitivity & \B 0.808 & 0.722 & 0.758 & 0.747 & \B 0.808 \\
    & Specificity & 0.633 & 0.507 & 0.702 & 0.651 & 0.629 \\
    & Precision   & 0.751 & 0.671 & 0.777 & 0.744 & 0.751 \\
    & F1          & 0.773 & 0.688 & 0.764 & 0.743 & 0.773 \\
    & VMS         & 0.205 &	0.276 &	0.167 & 0.138 & 0.161 \\
    \midrule
    \multirow{7}{\linewidth}{JF} 
    & AUROC       & 0.784 & 0.610 & 0.769 & 0.790 & 0.781 \\
    & Accuracy    & 0.714 & 0.636 & 0.708 & 0.728 & 0.697 \\
    & Sensitivity & 0.792 & 0.658 & 0.698 & 0.732 & 0.683 \\
    & Specificity & 0.611 & 0.607 & 0.720 & 0.722 & 0.716 \\
    & Precision   & 0.737 & 0.695 & 0.780 & 0.780 & 0.763 \\
    & F1          & 0.755 & 0.674 & 0.731 & 0.753 & 0.719 \\
    & VMS         & 0.230 &	0.169 &	0.209 &	0.117 & 0.109 \\
    \bottomrule
    \multicolumn{7}{p{250pt}}{AUROC: area under the receiver operating characteristics curve. EF: early fusion; JF: joint fusion; LF: late fusion.}\\
\end{tabularx}
\end{table}


\begin{table}
\centering
    \caption{Performance for different modality combinations using CatBoost}
    \label{table_catboost}
    \begin{tabularx}{\columnwidth}{p{30pt}>{\centering\arraybackslash}p{35pt}>{\centering\arraybackslash}p{35pt}>{\centering\arraybackslash}p{45pt}>{\centering\arraybackslash}p{55pt}}
    \toprule
    & Tabular \break + Text (EF) & Tabular \break + Text (LF) & Tabular + Audio (LF) & Tabular + Text + Audio (Mix)\\
    \midrule
    AUROC          & \B 0.829 & 0.824 & 0.669 & 0.728 \\
    Accuracy     & \B 0.810 & 0.752 & 0.646 & 0.669 \\
    Sensitivity  & 0.792 & 0.817 & 0.682 & 0.733 \\
    Specificity  & \B 0.833 & 0.667 & 0.598 & 0.582 \\
    Precision    & \B 0.867 & 0.768 & 0.697 & 0.707 \\
    F1           & \B 0.792 & 0.788 & 0.686 & 0.713 \\
    VMS          & \B 0.065 & 0.174 & 0.138 & 0.203 \\
    \bottomrule
    \multicolumn{5}{p{240pt}}{EF: early fusion; LF: late fusion; Mix: mixture combination. Tabular + text with CatBoost using EF, audio was trained with nural network, the two models were combined in LF; AUROC: area under the receiver operating characteristics curve.}\\
    \end{tabularx}
\end{table}


As generalization in all aspects is considered more important than skewed performance in prediction models, we consider an outperform of higher number of metrics than others as a better approach. The unimodal prediction results of tabular and text data were compared (shown in Table~\ref{table_tree_tabular} and \ref{table_tree_text} in Appendix). CatBoost achieved the best performed tree-based algorithm for tabular data, and 
NN outperform others in text data. STFT spectrogram outperformed MFCC (shown in Table~\ref{table_uni_audio} in Appendix), therefore, were used in the following experiments. Table~\ref{table_singular} summarized the results for best performed unimodals. Free text outperformed other modalities, achieving AUROC and accuracy of 0.85 and 0.76, respectively. 

In the multimodal design, Table~\ref{table_EF_JF} and Table~\ref{table_catboost} demonstrate the results of applying NN and CatBoost using different fusion strategies. Tabular + text using CatBoost in EF achieved the best performance among all combinations. In two-modality design, EF and LF commonly suffered from low specificity, adding audio seems to improve specificity. JF appears to be a better approach to fuse audio modality. In triple-modality design, JF demonstrated generally well performances with less VMS, and was considered to be a more robust approach to combine triple modalities. Hybrid approach EF + LF outperformed EF, but JF + LF did not performed better than JF. A mixture combination was additionally tested based on the results. Concatenating tabular + text with CatBoost in EF and integrated audio features that trained with NN in LF, shown in Table~\ref{table_catboost}. However, the performance of the mixture version did not standout. 


In $r$ analysis between tabular features and the prediction target (shown in Fig.~\ref{fig_correlation} in Appendix), VAS was the highest linear correlated feature with prognosis at 0.24. However, it was not higher than ±0.70, implying that the variables were insignificant and unlikely to dominate the prediction. In CCA analysis, high correlated canonical variates between all three modalities ($r$ = 1.00) was found pair-wisely. Nevertheless, only text and audio were able to achieve high correlated canonical variates between the prediction target ($r$ = 1.00), respectively. The highest $r$ between tabular data and the prediction targets were 0.47. According to the $weights$ and $loadings$ in CCA (demonstrated in Table~\ref{table_CCA} in Appendix), one unit increase in ODI leads to 0.49 increment in text canonical variate ($weights$ = 0.49) when all of the other variables are held constant. Similarly, \emph{Stasis score} lead to 0.65 increment in audio canonical variate ($weights$ = 0.65). Age correlated and contributed the most to the text canonical variate ($loadings$ = 0.88), followed by ASA ($loadings$ = 0.79); and BMI contributed the most to audio canonical variate ($loadings$ = 0.82), with \emph{Yang-Xu score} contributed secondly ($loadings$ = -0.62). The negative sign indicates the affecting direction, and the value indicates the affecting volume.

\begin{figure}
\centering
    (a){\includegraphics[width=2in]{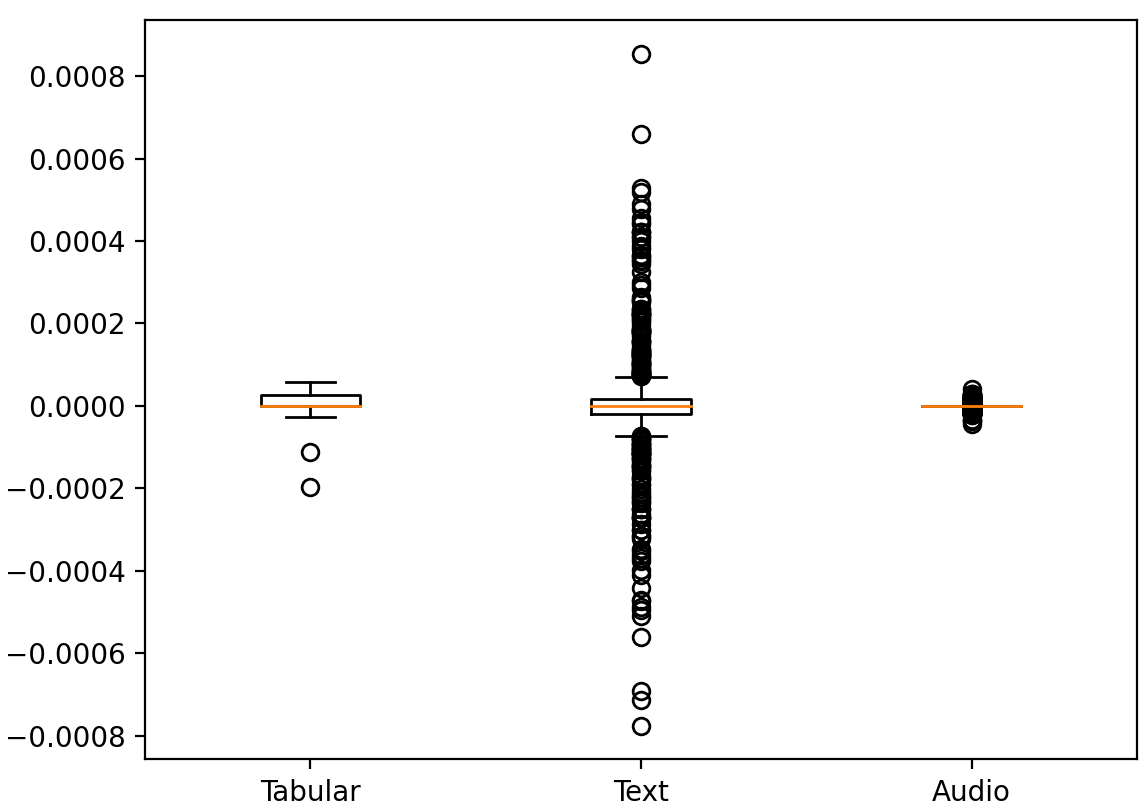}}
\centering
    (b){\includegraphics[width=2in]{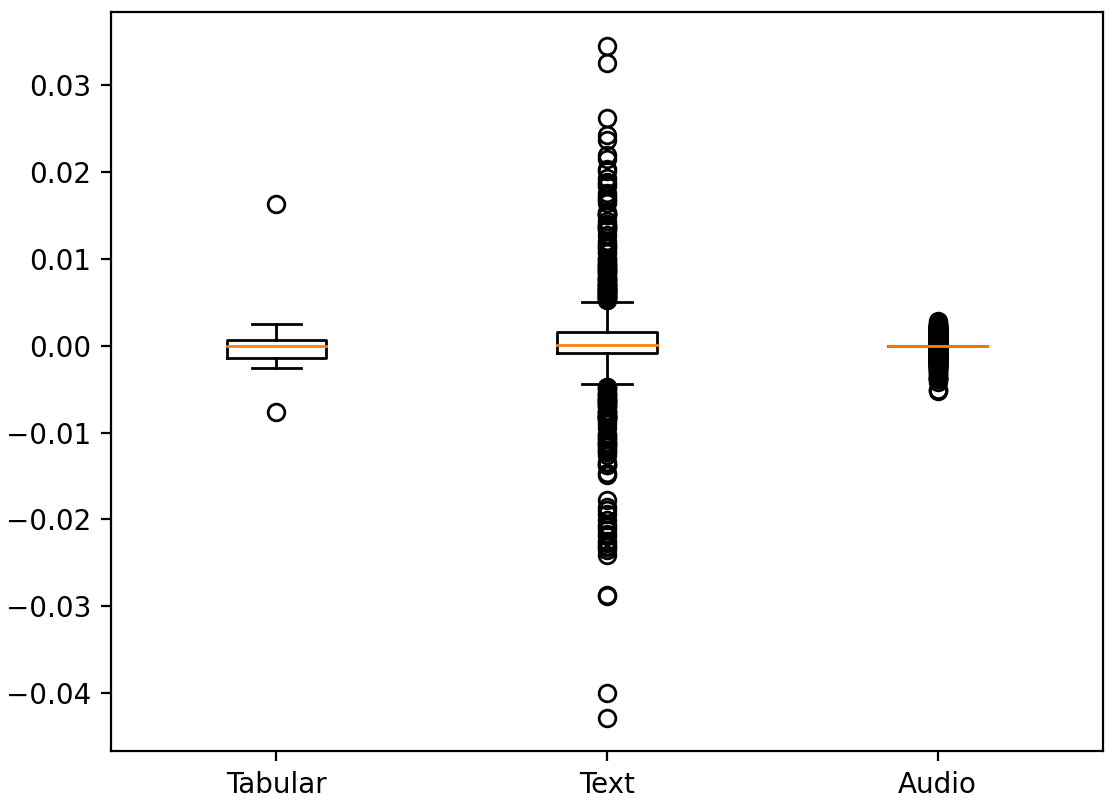}}
\caption{Integrated Gradients analysis for interpretability of the (a) EF and (b) JF model. The vertical axis represents the value distribution of the $attribute$ of each modality. $attribute$: the contribution of the input to the prediction results.}
\label{fig_IG}
\end{figure}

In IG analysis, Fig.~\ref{fig_IG} demonstrates the distribution of the $attribute$ of (a) EF and (b) JF model in triple modalities combination. The two models shared the same trend, the values were relatively small and centralized for tabular and audio data, whereas the text data having much larger values and wider variations during the prediction. It is worth noticing that the values for JF were significantly larger than EF. In the analysis of CatBoost feature importance in unimodal tabular modality training (shown in Fig.~\ref{fig_feature_importance} in Appendix), age (32.38), EQ-5D (15.46), and ASA (11.02) were listed as the features having the highest importance. Further, when we analyzed the model trained with tabular + text (using CatBoost in EF), the sum of the importance for tabular is zero, leaving text with 100.  

\begin{figure}
\centerline{\includegraphics[width=0.8\columnwidth]{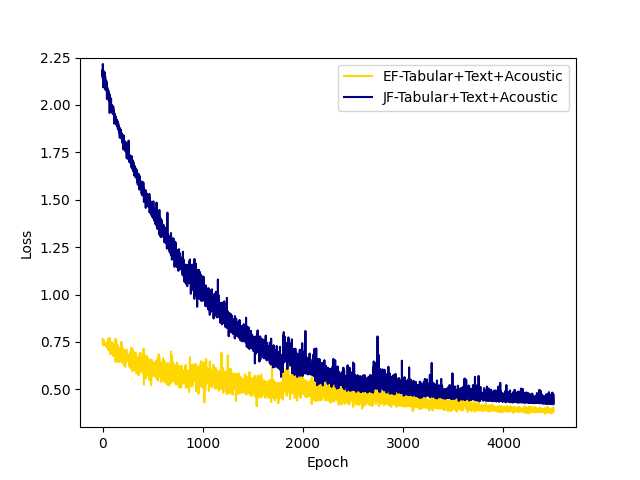}}
\caption{Trend of the condensing loss during training for triple multimodal combinations in EF and JF design. EF: early fusion; JF: join fusion.}
\label{fig_tri_EF_JF}
\end{figure}

Fig.~\ref{fig_tri_EF_JF} shows the trend of the condensing loss during training for triple multimodal combinations in EF and JF. It can be observed that JF reached a minimum loss gradually, whereas EF did not condensed as much, arriving at its' minimum quickly. The condensing trend for two modalities followed the same pattern.

\section{DISCUSSION} 
Our study explored the potential of using preoperative assessments to evaluate the prognosis of patients before lumbar spinal surgery. We have demonstrated the usage of multi-modalities, multimodals, and different fusion strategies can be beneficial. To the best of our knowledge, this is the first study to demonstrate the possibility of using machine learning in prognosis evaluation of lumbar spinal surgery. Our light-weighted instrument merely requires questionnaire assessments, which is highly achievable in general practice. 

\subsection{Modality differences} 
Meanwhile, our study provides insights into leveraging multiple combination of modalities, multimodals, and different fusion strategies. In our study, merely using the description of how the surgery was planned provided a significant amount of information that corresponded to the prognosis, enabling high prediction accuracy. This is consistent with the medical perspective that the surgical approaches and spinal segments significantly influence the outcome of the surgery \cite{Tulder2005OutcomeOI,Jacobs2012SurgicalTF}. The description remained in the form of free text because the surgery was tailored according to each patient’s condition although the surgical approaches were approximately the same. It is more informative than categorizing the context into selectable items. However, such information has traditionally been difficult to capture because of potential errors, misspellings, and human interpretable language, which cannot be used without sophisticated preprocessing. NLP pre-trained technology facilitates the extraction of meaningful embeddings and enables applications to make use of the information easily. The results of IG analysis and feature importance analysis suggested the same information that the text features contributed more than other modalities. The superior result obtained in text features can also be attributed by appropriate representation transformation \cite{huang2020fusion}, enabling the following classifier to make appropriate predictions. 

Tabular data consists of information that is most commonly used in existing practice. The variables were well-known indicators accumulated from medical research. The pre-post differentiation of assessments are the basic materials that define the prognosis of the patient. Nevertheless, according to the results of $r$ and CCA, linear correlations between prognosis and assessments cannot be established with merely information from pre-assessments, exhibiting the limitation in current practice. CatBoost is able to capture the underlying patterns between prognosis and pre-assessments, and performed better than NN. Although not all of our inputs are categorical variables, but the task does required to find a threshold for discriminating one category from another. This type of task can be difficult for NN, but could be naturally handled by CatBoost. This is also true when concatenating tabular and text features in EF, which achieved the best performance among all combinations. Another reason may be the requirement of samples. Different algorithms requires different sample size to achieve a stable performance \cite{adequate_sample}. Tree-based algorithms requires less samples than NN. Nevertheless, a relatively small volume of samples is common in medical studies. Thereby, our work corresponded to real world scenario. In our research, CatBoost and its' combinations demonstrate lower variability, which was previously mentioned \cite{NN_Tree} that when a decision tree had lower error than NN on a dataset, then the tree-based ensemble methods are tend to have lower error than NN as well. 

Although the audio recordings did not contributed as expected. It act as an auxiliary and improved the generalizability and harmonization of the model under proper fusion strategies, implying that more information does improve model performance. The vowel pronunciations indicates the inner energy imbalance of an individual \cite{lin2012bcqs, chiu2000objective} in Eastern medicine. The energy was the foundation of TCM BCQ body constitution and implies the reaction of individuals towards diseases and treatments \cite{Zhu2017AssociationBN,Liang2020ClinicalRL, You2017AssociationOT,Dong2021AnalysesOL}. However, the referred "energy" was a concept that was not specifically related to acoustic features, which were difficult to capture.

\subsection{Fusion strategy differences}
Several results in our work indicates the different nature of EF and JF. EF is suitable for features that were readily processed for classification. In our work, the tabular variables were well-known medical indicators, and text features were processed by pretrained PubMedBERT, both were distinguishable features. Tabular + text using CatBoost and NN in EF both outperform JF, indicating well processed features do not require to converge as much. Whereas JF valued each modality more than EF (having $attribute$ larger and wider in distribution) and converge more than EF. JF is suitable for features that required further extraction, such as the audio modality in our work was not processed with a well defined pre-trained model. With more sophisticated feature extraction, audio modality was better integrated and contributed in JF. Different modalities possessed diverse information that requires different feature extraction to manifest its’ value, and the join-trained classification layers learned to do more harmonized integration. JF achieved the best performance in triple-modality design, which is a more generalized model ensuring generally high performances in all aspects with less variation. In our study, the variation of the models achieved minimum in triple-modality design.


The learning behavior of LF was difficult to observe due to the final fusion was not based on learned parameters. Although LF can achieve high AUROC and accuracy, they suffered from low specificity, it may requires a grid search to find the perfect way of combining the results. In our experiences, the fusion approach in multimodals is better done through learning. Hybrid version of heterogeneous machine learning methods and a mixture of fusion strategies does not necessarily lead to better results.  Meanwhile, although the tabular features, or the concatenated tabular, text, and audio features do not correspond to spatial or temporal structure, in our experiment, treating them in an image-like approach and used a CNN-based encoder performed better than FC layers. This phenomenon was also shown previously \cite{diabetes_JBHI,CNN_tabular}, which symbolized that static and concatenated vectors can be representative as a general status of patients and made CNN-based encoders effective.

\subsection{Interrelationship among modalities}
The correlation between the three modalities, and their correlation between surgical outcomes had seldom been explored. In our study, the inner correlation between triple modalities were high, which is reasonable for the information in different modalities are gathered from the same patient. Age and ASA are factors that were repeatedly indicated in CCA and feature importance analysis, symbolizing it is closely connected with the surgical plans. Elder patients are at higher risk of being diagnosed with LBP and sciatica, suggesting longer duration \cite{cook2014risk}, slower self-healing abilities, and other chronic diseases. ASA is a known indicator to evaluate the risks based on the physiological status of patients. Elderly with higher complications naturally increased the risk of ASA. BMI and \emph{Yang-Xu score} both indicate the composition of the body, which potentially signals that the vowel pronunciations are related to individual differences \cite{chiu2000objective}.

The number of patients recruited was relatively small. However, based on the sample size estimation method proposed by \cite{Riley_sample_size_estimation} and \cite{sample_size_estimation2}, our work falls in the absolute margin of error between 0.05 and 0.1 under 95\% CI (requiring 100 to 408 participants). This indicates the credibility of our work fall at the confidence of 95\%. Therefore, we considered our sample size sufficient. The pathological mechanisms of using TCM body constitution and vowel pronunciation to evaluate lumbar spinal surgery prognosis have yet to be widely proven in Western medicine. However, evidence was found indicating vowel pronunciation is associated with hypertension, coronary artery diseases, and several chronic diseases \cite{chiu2000objective, maor2018voice}, and the stability of circulation and heart rate variability would affect spinal surgery outcomes \cite{sodervall2013heart}. Thereby, more discovery in the relationship of vowel pronunciations, circulation problems, and spinal surgeries may be promising.


\section{CONCLUSION}
Through adding the elements of Eastern medicine and machine learning technique, we successfully developed an effective tool to foreseeing surgical prognosis of lumbar spinal surgery using multimodalities and multimodal. Our approach can be widely applied in general practice due to simplicity and effective. We provided insights into leveraging multiple combinations of modalities, multimodals, and fusion strategies. Tree-based algorithms can be applied when the data possessed low dimensionality, smaller sample size, and the task required discriminating thresholds for classification; NN layers are better at data with high dimensionality or required feature extraction, and performed better integration with heterogeneous modalities. Among different fusion strategies, EF is suitable for readily processed features; JF integrated different modalities harmoniously; LF integrated heterogeneous machine learning methods. Machine learning has the potential to overcome the insufficiency of current practices. Further investigation in applying this application in practices and observing its' effects are promising.


\pagebreak
\section*{Appendix}
{\appendices
\begin{table}[!ht]
\caption{Prognosis determination information of the recruit patients}
\label{appendix_table_demographic}
\setlength{\tabcolsep}{3pt}
\begin{tabular}{p{55pt}p{140pt}p{30pt}p{25pt}}
\toprule
\multicolumn{2}{c}{Items} & \multicolumn{2}{c}{Values}\\
\midrule  
\multirow{5}{\linewidth}{Patient Diagnosis} 
          & Cervical Spondylosis, n (\%)         & 3   &	0.029 \\
          & Degenerative kyphoscoliosis, n (\%)  & 1   &	0.010 \\
          & Herniated Intervertebral Disc, n (\%)&	26 &	0.248 \\
          & Infectious Spondylodiscitis, n (\%)  &	1  &	0.010 \\
          & Spondylolisthesis, n (\%)            &	71 &	0.676 \\
          & Kyphoscoliosis, n (\%)               &  3  &    0.029 \\
\midrule  
\multirow{6}{\linewidth}{Surgical Approach} 
          & Anterior Cervical Discectomy and Fusion, n (\%)                          & 1  & 0.010 \\
          & Anterior Cervical Corpectomy with Fusion (ACCF), n (\%)                  & 2  & 0.019\\
          & Minimally Invasive Surgery Transforaminal Lumbar Interbody Fusion, n (\%)& 73 &	0.695 \\
          & Percutaneous Endoscopic Discectomy and Drainage, n (\%)	                 & 1  & 0.010 \\
          & Percutaneous Endoscopic Discectomy and Foraminoplasty, n (\%)            & 1  & 0.010 \\
          & Percutaneous Endoscopic Lumbar disc Discotomy, n (\%)                    & 14 &	0.133 \\
          & Percutaneous Endoscopic Lumbar Laminectomy and Discectomy, n (\%)        & 12 & 0.114\\
          & Transforaminal Lumbar Interbody Fusion, n (\%)                           & 1  & 0.010 \\
\midrule  
\multirow{6}{\linewidth}{Operative Assessment (postoperative)} 
          &VAS (mean, SD)&	2.648 & 1.813\\
          &VAS pre-post differentiation (mean, SD)&	-1.724&	2.759\\
          &EQ-5D (mean, SD)	&0.635&	0.117\\
          &EQ-5D pre-post differentiation (mean, SD)&	0.023 & 0.158\\
          &ODI (mean, SD)&	0.456 &	0.171\\
          &ODI pre-post differentiation (mean, SD)&	-1.691 & 5.395\\
\midrule 
\multirow{5}{\linewidth}{Surgery and Administration Status} 
          &Total surgery time (minute) (mean, SD)&	231.124 & 93.406\\
          &Amount of blood loss (mean, SD)&	94.238&	100.713\\
          &Types of analgesic drugs used (mean, SD)&	3.333&	1.198\\
          &Total admission days (mean, SD)&	5.638&	3.235\\
          &Complications occurred during the administration, n (\%) & 7 & 0.067\\
\midrule
\multirow{1}{\linewidth}{Prognosis Determination} 
          &Number of desirable outcome labels (mean, SD)&	4.771 & 2.027\\
          
\bottomrule
\multicolumn{4}{p{270pt}}{n: number of samples; SD: Standard deviation; VAS: Visual Analog Scale; EQ-5D: EuroQol Five Dimensions; ODI: Oswestry Disability Index; pre-post differentiation: $|pre-post|/pre$.} \\
\end{tabular}
\end{table}

\begin{table}
\centering
    \caption{Unimodals performing on tabular data.}
    \label{table_tree_tabular}
    \begin{tabularx}{\columnwidth}{p{30pt}>{\centering\arraybackslash}p{25pt}>{\centering\arraybackslash}p{25pt}>{\centering\arraybackslash}p{25pt}>{\centering\arraybackslash}p{25pt}>{\centering\arraybackslash}p{25pt}>{\centering\arraybackslash}p{25pt}}
    \toprule
    & RF & Ada & Cat & LGBM & XGB & NN\\
    \midrule
    AUC         & 0.713 & 0.773 & \B 0.815 & 0.769 & 0.681 & 0.670 \\
    Accuracy    & 0.714 & 0.714 & \B 0.738 & 0.667 & 0.714 & 0.667   \\
    Sensitivity & 0.750 & 0.750 & 0.792 & \B 1.000 & 0.917 & 0.567   \\
    Specificity & 0.667 & 0.667 & 0.667 & 0.222 & 0.444 & \B 0.800 \\
    Precision   & 0.750 & 0.750 & 0.760 & 0.632 & 0.688 & \B 0.797   \\
    F1          & 0.750 & 0.750 & \B 0.775 & 0.774 & 0.786 & 0.658   \\
    \bottomrule
    \multicolumn{7}{p{250pt}}{RF: Random Forest; Ada: Adaptive Boosting; Cat: Categorical Boosting; XGB: eXtreme Gradient Boosting; LGBM: Light Gradient Boosting Machine; NN: Neural Network; AUROC: area under the receiver operating characteristics curve.}\\
    \end{tabularx}
\end{table}

\begin{table}
\centering
    \caption{Unimodals performing on text data.}
    \label{table_tree_text}
    \begin{tabularx}{\columnwidth}{p{30pt}>{\centering\arraybackslash}p{25pt}>{\centering\arraybackslash}p{25pt}>{\centering\arraybackslash}p{25pt}>{\centering\arraybackslash}p{25pt}>{\centering\arraybackslash}p{25pt}>{\centering\arraybackslash}p{25pt}}
    \toprule
    & RF & Ada & Cat & LGBM & XGB & NN\\
    \midrule
    AUC         & 0.787 & 0.778 & 0.824 & 0.685 & 0.597 & \B 0.845 \\
    Accuracy    & \B 0.762 & 0.619 & 0.714 & 0.571 & 0.619 & \B 0.762 \\
    Sensitivity & 0.833 & 0.917 & \B 0.875 & 0.667 & 0.750 & 0.850\\
    Specificity & \B 0.667 & 0.222 & 0.500 & 0.444 & 0.444 & 0.644\\
    Precision   & \B 0.769 & 0.611 & 0.701 & 0.615 & 0.643 & 0.768\\
    F1          & 0.800 & 0.733 & \B 0.875 & 0.640 & 0.692 & 0.801 \\
    \bottomrule
    \multicolumn{7}{p{250pt}}{RF: Random Forest; Ada: Adaptive Boosting; Cat: Categorical Boosting; XGB: eXtreme Gradient Boosting; LGBM: Light Gradient Boosting Machine; NN: Neural Network; AUROC: area under the receiver operating characteristics curve. NN outperfromed CatBoost in 4 metrics, and outperformed Random Forest in metrics average (0.770 and 0.778).}\\
    \end{tabularx}
\end{table}


\begin{table}
\centering
    \caption{Unimodal performing with difference acoustic features.}
    \label{table_uni_audio}
    \begin{tabularx}{\columnwidth}{p{30pt}>{\centering\arraybackslash}p{50pt}>{\centering\arraybackslash}p{50pt}}
    \toprule
    & STFT & MFCC \\
    \midrule
    AUC         & 0.599 & 0.594 \\
    Accuracy    & 0.607 & 0.583 \\
    Sensitivity & 0.613 & 0.650 \\
    Specificity & 0.598 & 0.493 \\
    Precision   & 0.672 & 0.637 \\
    F1          & 0.640 & 0.639 \\
    \bottomrule
    \multicolumn{3}{p{250pt}}{STFT: Short-time Fourier transform spectrogram; MFCC: Mel-frequency cepstrum.}\\
    \end{tabularx}
\end{table}

\pagebreak
\begin{figure}
\centerline{\includegraphics[width=\columnwidth]{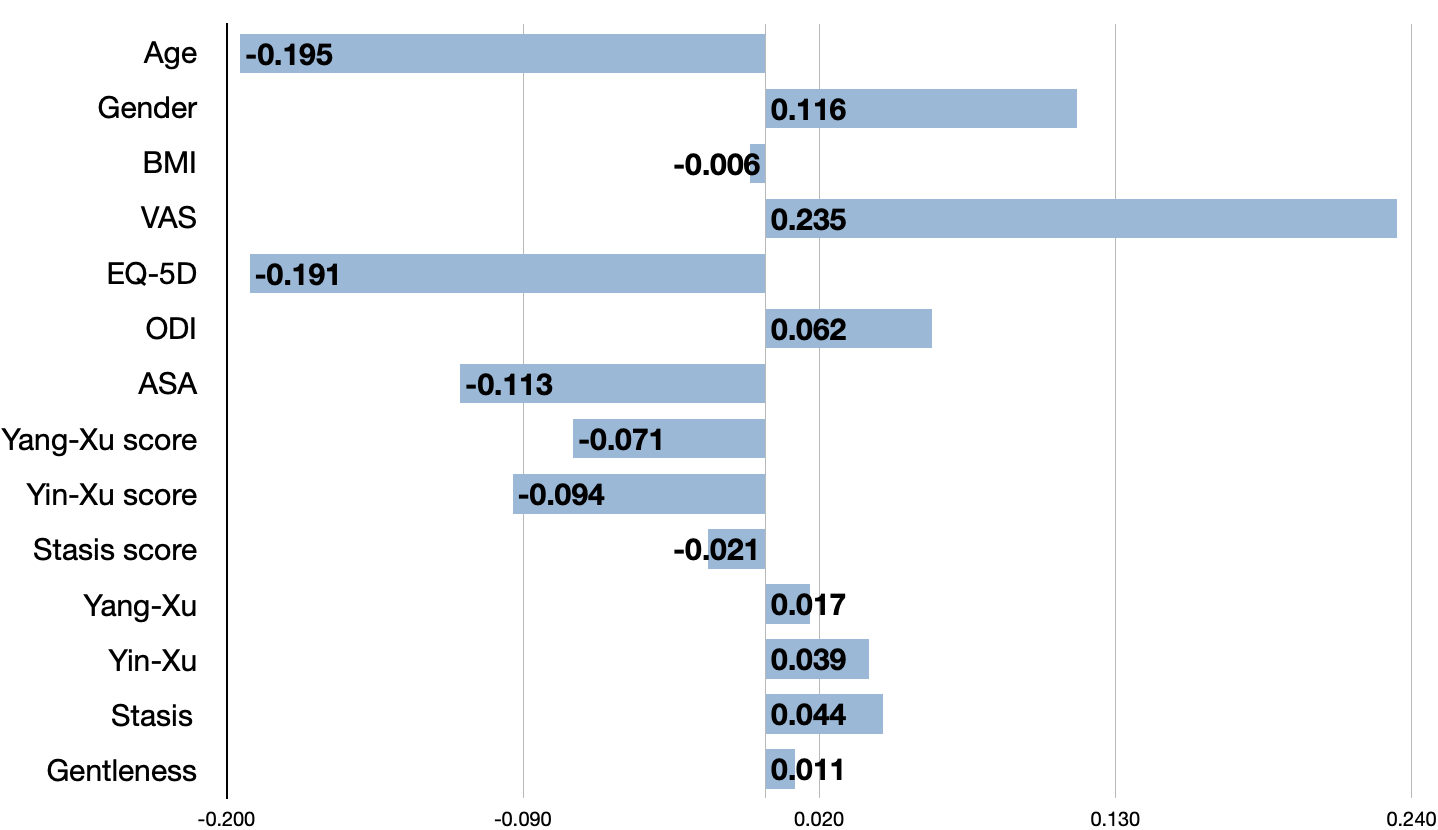}}
\caption{Pearson correlation test ($r$) of the tabular features against the prediction target. BMI: Body mass index; VAS: Visual Analog Scale; EQ-5D: EuroQol Five Dimensions; ODI: Oswestry Disability Index; ASA: American Society of Anesthesiologists.}
\label{fig_correlation}
\end{figure}


\begin{table}
\centering
    \caption{Canonical correlation analysis.}
    \label{table_CCA}
    \begin{tabularx}{\columnwidth}{p{35pt}>{\centering\arraybackslash}p{37pt}>{\centering\arraybackslash}p{37pt}>{\centering\arraybackslash}p{37pt}>{\centering\arraybackslash}p{37pt}}
    \toprule
    \multirow{3}{\linewidth}{} 
        & \multicolumn{2}{c}{Tabular correlated to text}
        & \multicolumn{2}{c}{Tabular correlated to audio}\\
    \cmidrule(lr){2-3} \cmidrule(lr){4-5}    
    & $weights$ & $loadings$ & $weights$ & $loadings$ \\
    \midrule
    Age           & 0.165  & \B 0.878  & -0.027 & -0.207 \\
    Gender        & -0.177 & -0.497 & -0.138 & -0.431 \\
    BMI           & 0.221  & 0.440  & 0.228  & \B 0.820  \\
    VAS           & -0.358 & 0.442  & -0.130 & -1.244 \\
    EQ-5D         & 0.419  & 0.157  & 0.194  & 1.270  \\
    ODI           & \B 0.494  & 0.333  & 0.076  & -1.214 \\
    ASA           & 0.206  & 0.785  & 0.173  & 0.199  \\
    \emph{Yang-Xu score} & -0.266 & -0.063 & -0.467 & -0.710 \\
    \emph{Yin-Xu score}  & 0.188  & 0.135  & -0.311 & -0.621 \\
    \emph{Stasis score}  & 0.306  & -0.017 & \B 0.649  & -0.482 \\
    \emph{Yang-Xu}       & -0.152 & -0.273 & 0.274  & -0.223 \\
    \emph{Yin-Xu}        & -0.205 & -0.258 & 0.045  & -0.485 \\
    \emph{Stasis}        & -0.087 & -0.233 & -0.059 & -0.518 \\
    \emph{Gentleness}    & -0.173 & 0.191  & 0.143  & 0.368  \\
    \bottomrule
    \multicolumn{5}{p{250pt}}{$weights$: analogous to coefficients of regressions, indicating the amount of one unit increase in variable leads to variance in canonical variate when all other variable held constant; $loadings$: correlation coefficients, referred to contribution and correlation to the canonical variate.}\\
    \end{tabularx}
\end{table}

\pagebreak
\begin{figure}[!ht]
\centerline{\includegraphics[width=\columnwidth]{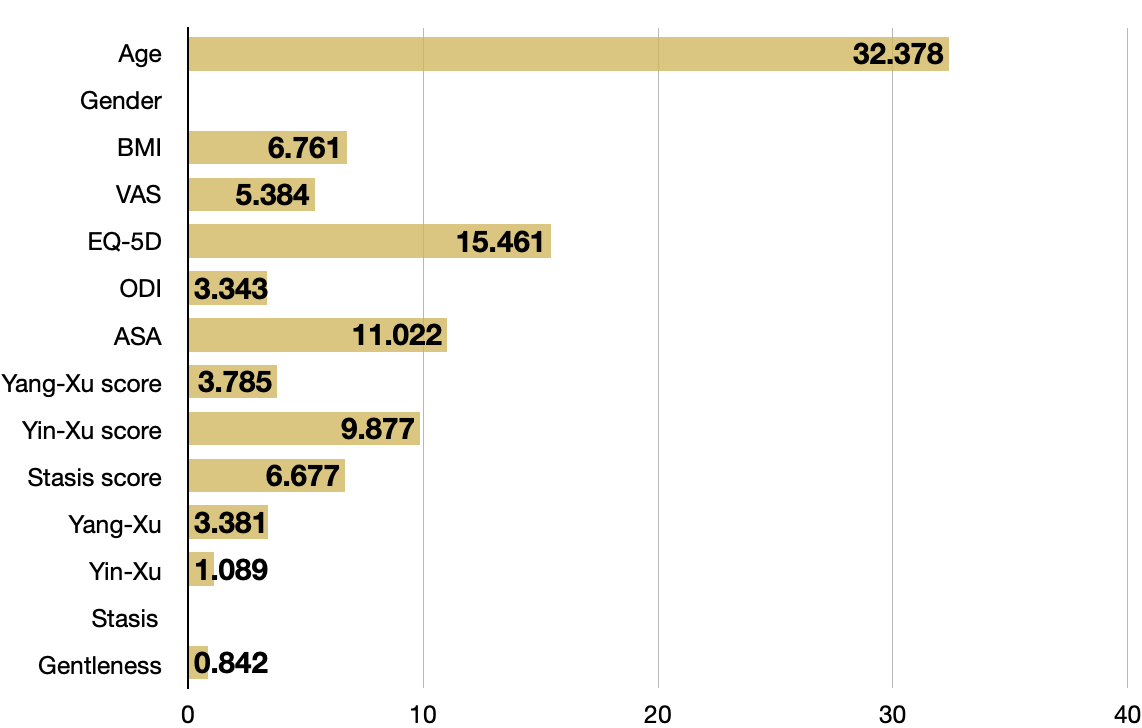}}
\caption{Feature importance values are normalized that the sum of importances of all features is equal to 100. BMI: Body mass index; VAS: Visual Analog Scale; EQ-5D: EuroQol Five Dimensions; ODI: Oswestry Disability Index; ASA: American Society of Anesthesiologists.}
\label{fig_feature_importance}
\end{figure}

}

\bibliographystyle{IEEEtran}
\bibliography{refs}	

\end{document}